# Structural Tags, Annealing and Automatic Word Classification

J. McMahon* and F.J. Smith
The Queen's University of Belfast

May 28, 1994

**Abstract**

This paper describes an automatic word classification system which uses a locally optimal annealing algorithm and average class mutual information. A new word-class representation, the *structural tag* is introduced and its advantages for use in statistical language modelling are presented. A summary of some results with the one million word LOB corpus is given; the algorithm is also shown to discover the vowel-consonant distinction and displays an ability to cluster words syntactically in a Latin corpus. Finally, a comparison is made between the current classification system and several leading alternative systems, which shows that the current system performs tolerably well.

## 1 Introduction

This paper contains a description of some work on an automatic word classification system which uses a technique similar to annealing [1]. The automatic acquisition of word classes corresponds to the paradagmatic component [5] of the syntagmatic-paradagmatic bootstrapping problem [19]. The best of the recent classification algorithms come in various forms [12, 8, 17, 2, 6, 3, 23] but most share underlying similarities which can be expressed best in the language of information theory [24, 14].

Over three hundred years ago, the Right Reverend John Wilkins presented his ideas about a *universal character* to the Royal Society [25] : this universal character was an artificial language where the structure of the words stood in a supposedly logical and universal relationship to objects in the world; this vision has been shared by many other language scholars — *e.g.*Bacon, Dalgarno,

*Supported by the Department of Education for Northern Ireland and British Telecom Research Labs. VODIS supplied by B.T., all other corpora by the Oxford Text Archive. Contact address, Department of Computer Science, Q.U.B., Belfast BT7 1NN, N. Ireland.

Lodwick, Leibniz, Comenius, Frege, Peano, Russell and Wittgenstein. Wilkins hoped that propositions, in his interlingua, would be "philosophically unfolded" and that pompous-sounding expressions should be summarily debunked; like Bacon before him and Ogden after him, he hankered after clear expression through the medium of a transparent language. He considered the redundancy of language as a design challenge rather than a necessary feature; he disapproved of equivocal words and synonyms and baulked at the ineffective design decisions of previous generations of language speakers. His system worked by dividing his experience of the world into classes and assigning (mostly) arbitrary consonants and vowels to the various classes and sub-classes. The word for a table might perhaps be "leda", where the first character represents a class of physical objects and the second represents the sub-classification of wooden objects, and so on. So the word for a desk might be the related word "ledu".

Strong arguments have been offered against the idea, from philosophical, linguistic and psychological perspectives (see [16] for a useful summarising discussion). Even his contemporary, Dalgarno criticised the detail of Wilkins' classification system, saying that poeple who spoke a foreign language would not agree with his rather culture-bound taxonomy. Nowadays, the arbitrariness of the linguistic sign is a tenet of modern linguistics; and the implicit reference theory of meaning has taken a philosophical battering. These criticisms notwithstanding, the Wilkins approach remains popular with the Artificial Intelligence community — reading Wilkins' chapter on 'the predicament of Quantity', which includes sub-divisions 'Of Magnitude' , 'Of Space' and 'Of Measure' is reminiscent of Hayes' *naïve physics manifesto* [11]; another chapter 'treats of action, and its several genus's 1. Spiritual 2. Corporeal 3. Motion 4. Operation'; here one is reminded of the work of Schank [22].

The main data structure used in the present work is the *structural tag*, an operationally defined analogue of a word from Wilkins' universal character. This way of representing words is not designed in order to be spoken by humans, nor to directly faciltate natural language translation, but to serve as a space within which words can be automatically clustered.

The end product of the classification process which will be described in the next section is a set of words, each of which is represented by a structural tag — a 16 bit (more generally, an $n$-bit) number whose binary representation specifies the location of that word in a cluster space. The structural tag corresponds to an easily accessible summary of the distributional properties of each word. The multi-modal nature of the distributions of some words — that is, the traditional linguistic problem of ambiguity — is as much a problem with this system as it is with others [7, 13], although theoretically, the structural tag should handle ambiguity: for example, the clustering performance of the algorithm described in this paper exhibits a differentiation between some unambiguous nouns and some lexical items which show verbal and noun distributions. Finch *et.al.* and Hughes *et.al.* both report similar clustering phenomena. The digital nature of the structural tag is not as serious a limitation as it initially appears to be [20].

With structural tags, classes can be conceived as *schemata* of the tag itself (using the standard genetic algorithm definition of schemata [10]). One advantage of thinking about the connection between words and classes in terms of bit patterns within structural tags, rather than as a distinct functional mapping between two distinct sets of objects, is that a no extra space is needed to store this class information; also, much less processing is required to derive class information, once the full structural tag is known. These considerations are important if one is interested in building statistical language models which will be using class-based information. Another advantage of using structural tags is that many levels of classification can be used simultaneously in the prediction of the probabilities of word segments; given that the acquisition of class information is so cheap using structural tags, this becomes a practical possibility in actual language model systems.

## 2  Word Clustering Method

Initially, a set of words is chosen to be clustered; these are usually the most frequent words of a given corpus, so that the unigram and bigram statistics which contain these words are more statistically significant — that is, their distributions in a corpus are reliable indicators of their distributions in natural language.

Each word is assigned a unique and random structural tag. This corresponds to a random, high entropy classification. The quality of the classification is measured by the average class mutual information [4],

$$M(f) = \sum_{c_i, c_j} P(c_i, c_j) \times \log \frac{P(c_i, c_j)}{P(c_i) P(c_j)} \qquad (1)$$

where $f$ represents some classification of these words. The classification algorithm works as follows (see figure 1): processing starts by concentrating on the first bit of every word — this corresponds to imagining that all words are classified as belonging to class 0 or class 1. This also corresponds to the most significant bit of the tag, and the coarsest possible grain of classification. Processing will not advance to the second bit of each word's tag representation until no word can be moved into its complementary class with a corresponding increase in average class mutual information. This is a locally optimal algorithm (of complexity $O(n^3)$); no globally optimal solutions exist at present.

Describing the algorithm informally, words flit around between different regions of the structural tag space, with tighter and tighter constraints on their movement as the bit processing moves from most significant to least significant. This is a type of simulated annealing process, the reverse of the usual bottom-up merge based clustering systems.

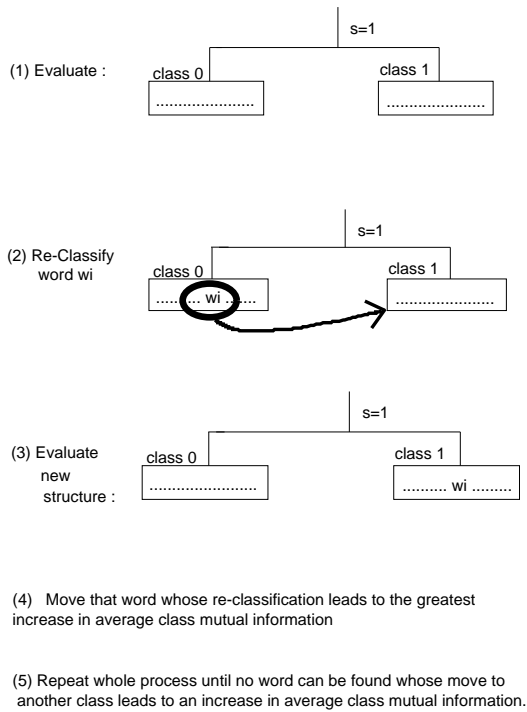

Figure 1: A classification, at depth 1, is evaluated. Local variations of it are also evaluated, by moving one word in turn into its complementary class. The best variation is chosen to be the new standard, whereupon the process is repeated until no variation is better than the standard classification.

# 3 Results

The million word LOB corpus was detagged, formatted and used to gather the raw word/class unigram and bigram frequency information. Syntactic and some semantic clustering emerged from the process. This result is summarised in figures 2, 3 and 4; figure 7 shows the overall topology of the structural tag classification space.

The smaller VODIS corpus in phonetic form was used to cluster phonemes in a similar way. This result is summarised in figure 5.

The assumption that this method is not specific to English is supported by the results obtained from a cluster of the complete works of Cicero, in Latin; these results are shown in figure 6.

The system compares tolerably well with some of the other word classification systems; Hughes [12] suggests an evaluation measure which estimates the degree of homogeneity of particular clusters within a classification. While not perfect [14], this is the only evaluation metric available at present. Figure 8 shows how the present system performs against two of the best word classifiers. It should be noted, however, that both of these comparison systems use contiguous and non-contiguous bigram information — that is, $\langle w_{x-2}, w_x \rangle$, $\langle w_{x-1}, w_x \rangle$, $\langle w_x, w_{x+1} \rangle$ and $\langle w_x, w_{x+2}, w_x \rangle$ bigrams. The system described in this paper only uses contiguous bigrams; some results in Hughes [12] suggest that the additional bigram information improves performance by approximately 3%.

A re-implementation of the merge-based approach described in Brown *et.al.* [3] and comparsion experiments, described in [14] identify the relative strengths and weaknesses of the two approaches; McMahon [14] also contains more results showing the strength of semantic clustering which can result from the most minimal definition of linguistic context possible — contiguous word bigrams; there is also a favourable comparison between the current system and an influential connectionist word clustering architecture described by Elman [6].

# 4 Conclusion

An annealing approach to automatic word classification, using average class mutual information as a metric produces linguistically interesting results. The structural tag representation facilitates this clustering and has several advantages if the resulting classification is to be used in statistical language modelling.

Many improvements could be made to the clustering algorithm; some of these are described in [14]. Further work on integrating a structural tag classification system into language models [15, 9] is currently being undertaken. No strong claims are made about the algorithm's psycholinguistic relevance, though we believe that the information processing paradigm upon which this research rests could be incorporated into either the traditional Chomskyan model of language acquisition [18] or its opposite [21].

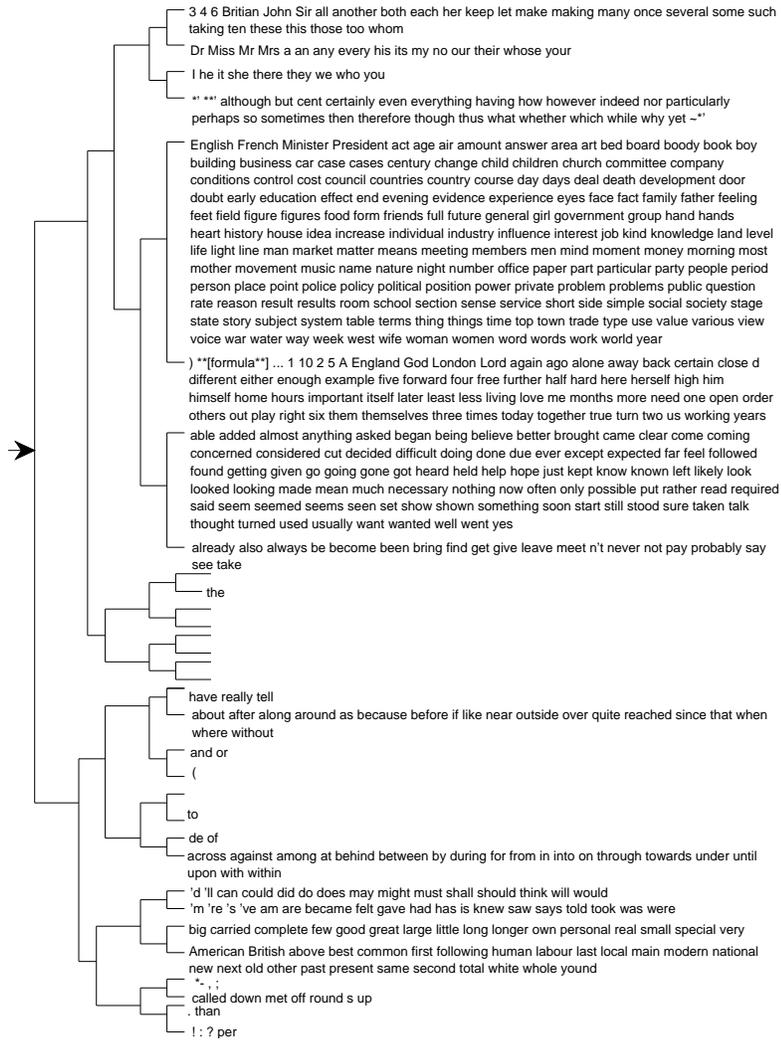

Figure 2: Final distribution of the most frequent words from the LOB corpus. Only the first five levels of classification are given here, but important syntactic relations are discovered.

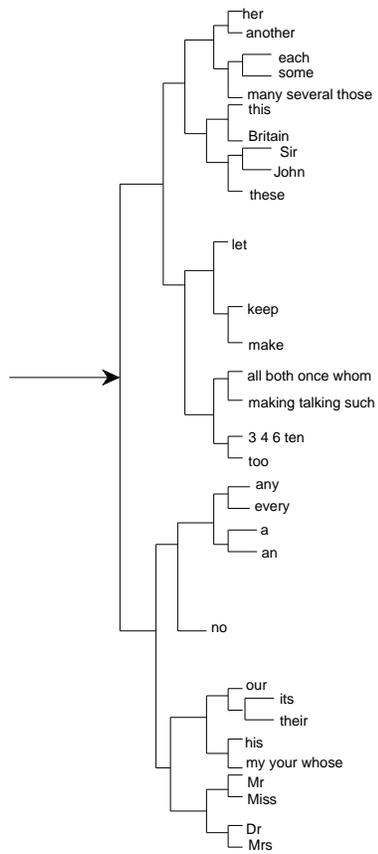

Figure 3: Detail of relationship between words whose final tag value starts with the four bits 0000. Many of these words exhibit determiner-like distributional behaviour.

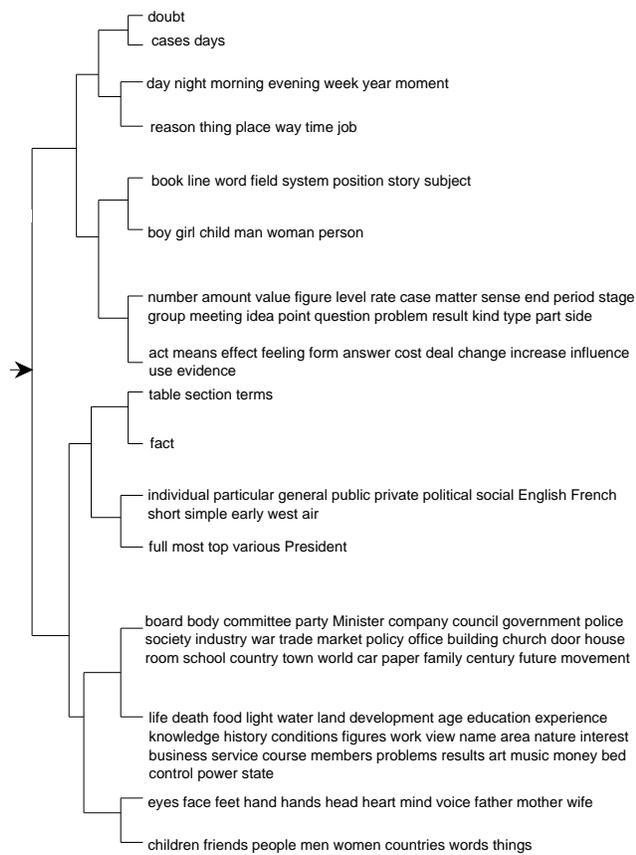

Figure 4: Detail, from level 5 to level 9, of many noun-like words. Clear semantic differences are registered.

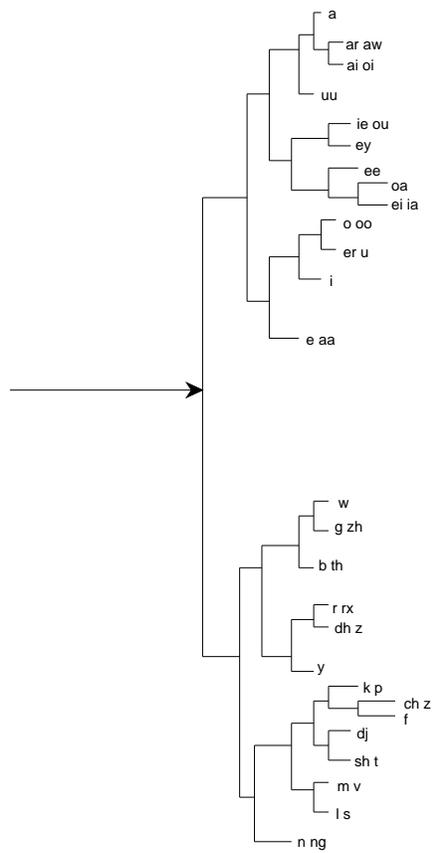

Figure 5: Automatic Phoneme Clustering which differentiates between vowels and consonants

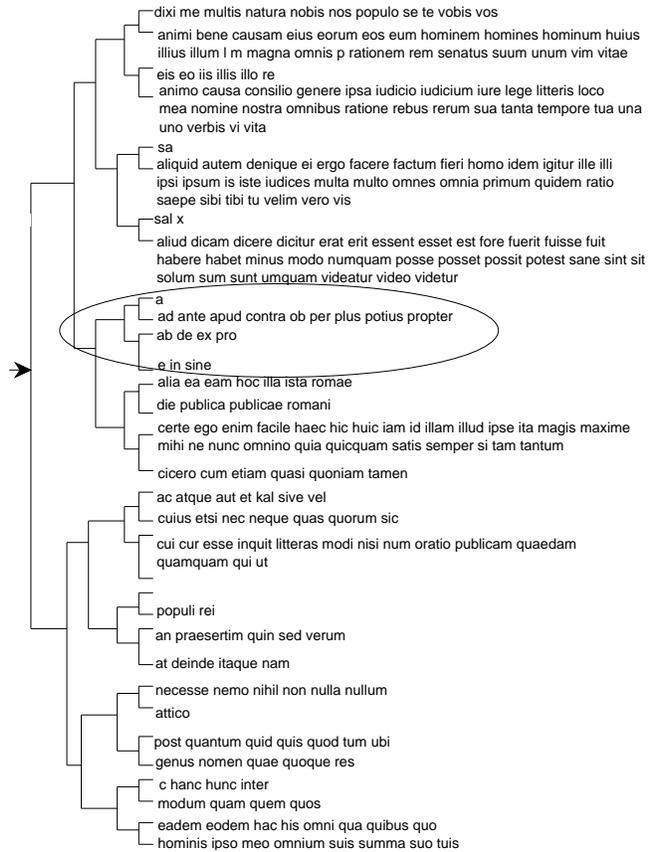

Figure 6: Classification of the most frequent words in a formatted version of the complete works of Cicero, in Latin; a group of prepositions is highlighted to show that the clustering system can find structure in languages other than English.

```
                ┌─ determiners - general
              ┌─┤  quantifiers
              │   determiners -possessive
            ┌─┤ pronouns - subject
            │ │ subordinating conjunctions
            │   WH-words
          ┌─┤─ nouns
          │ │
          │ │   nouns
          │ ├─ pronouns - object
          │ │   pronouns - reflexive
          │ │
          │ │   verbs - past participle
          │ ├─ verbs - past tense
          │ │   verbs - base form
          │ │
          │ └─   adjectives
        ►─┤ 'the'
          │   ┌─ adjectives
          │ ┌─┤─ adverbs
          │ │ └─ co-ordinating conjunctions
          ├─┤
          │ └─ prepositions
          │
          │ ┌─ auxiliaries
          ├─┤
          │ └─ adjectives
          │
          └─ punctuation
```

Figure 7: Approximate topology of the tree generated from the LOB corpus and the average class mutual information maximiser, using structural tags.

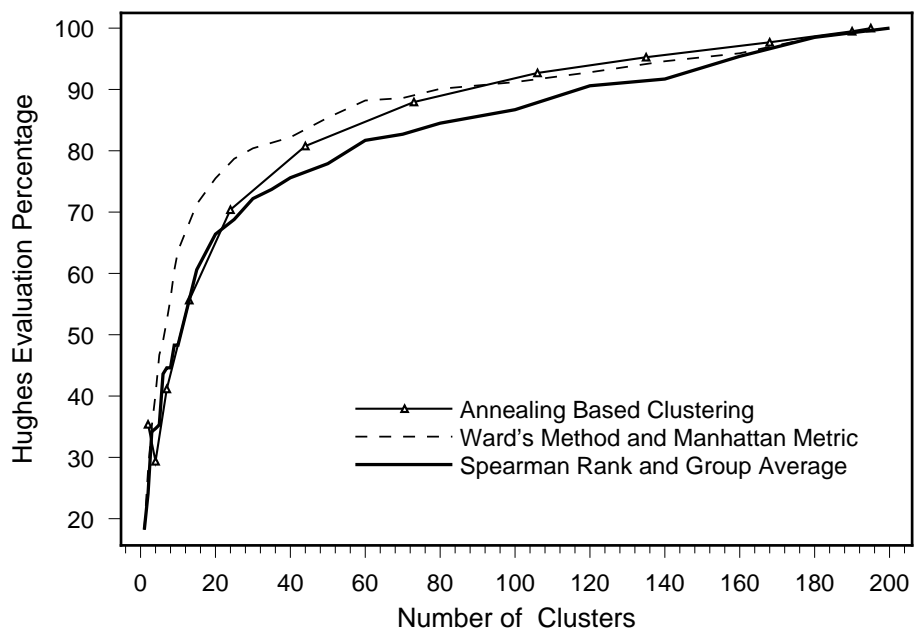

Figure 8: Graph showing the performance of the annealing classification system compared to two of the best of the current systems — those of Hughes and Atwell and Finch and Chater. Performance is measured by the Hughes-Atwell cluster evaluation system.